# Focus Group on Artificial Intelligence for Health

Marcel Salathé[1], Thomas Wiegand[2], Markus Wenzel[3]

Email: tsbfgai4h@itu.int
Web: https://www.itu.int/go/fgai4h

**Abstract**
Artificial Intelligence (AI) – the phenomenon of machines being able to solve problems that require human intelligence – has in the past decade seen an enormous rise of interest due to significant advances in effectiveness and use. The health sector, one of the most important sectors for societies and economies worldwide, is particularly interesting for AI applications, given the ongoing digitalisation of all types of health information. The potential for AI assistance in the health domain is immense, because AI can support medical decision making at reduced costs, everywhere. However, due to the complexity of AI algorithms, it is difficult to distinguish good from bad AI-based solutions and to understand their strengths and weaknesses, which is crucial for clarifying responsibilities and for building trust. For this reason, the International Telecommunication Union (ITU) has established a new Focus Group on "Artificial Intelligence for Health" (FG-AI4H) in partnership with the World Health Organization (WHO). Health and care services are usually the responsibility of a government - even when provided through private insurance systems - and thus under the responsibility of WHO/ITU member states. FG-AI4H will identify opportunities for international standardization, which will foster the application of AI to health issues on a global scale. In particular, it will establish a standardized assessment framework with open benchmarks for the evaluation of AI-based methods for health, such as AI-based diagnosis, triage or treatment decisions.

---

[1] Prof. Dr. Marcel Salathé is with EPFL (École Polytechnique Fédérale de Lausanne), Geneva, Switzerland
[2] Prof. Dr. Thomas Wiegand is with TU Berlin and Fraunhofer HHI, Berlin, Germany
[3] Dr. Markus Wenzel is with Fraunhofer HHI, Berlin, Germany

# Introduction

"The enjoyment of the highest attainable standard of health" is a basic human right (WHO Constitution, 1946). Good health for everyone has for centuries been a key goal of most governments, and public health breakthroughs such as vaccination are generally credited with having saved – and continuing to save – billions of lives. In many countries today, the healthcare industry is the largest and/or fastest growing industry, increasingly often accounting for more than ten percent of the gross domestic product. It is thus not surprising that the healthcare sector is a key area of application, when a technology reaches new levels of performance, as is the case with modern AI. Given the size of the health sector, the potential economic opportunities are immense. The potential for leveraging new technology for the common good by improving public health can also be enormous. It is therefore prudent to look at the potential of AI in helping solve health-related issues. This short paper describes the current applications of AI in the health domain and discuss challenges and how to address these in order to unlock the full potential of the technology.

# Artificial Intelligence

The term *artificial intelligence* is not new. As an academic field, it dates back to at least the mid-20$^{th}$ century, and has since gone through multiple cycles of substantial progress, followed by inflated expectation, and then disappointment. A combination of new machine learning algorithms, increased computational power, and an explosion in the availability of very large data sets ("big data"), as a consequence of the digitalisation of health information, has led to recent stunning advances, with demonstrations of machines achieving human-level competence at solving clearly defined tasks across many domains. The current cycle is primarily driven by the extremely impressive progress recently made by deep learning, a branch of machine learning that very effectively uses artificial neural networks to address harder problems than ever before. Applications of deep learning have achieved human or superhuman performance in many fields such as image recognition and natural language processing. Importantly, the neural network parameters are tuned in an automated process of iterative training. In many cases, no expert-level knowledge is used in the training process, other than direct input and output parameters (e.g. sets of pixels and their associated labels), giving rise to the so-called "end-to-end" learning. In other words, the networks learn to go directly from one end – the input – to the other end – the output – without requiring any domain-specific expertise in between. The resulting network structures are generally very large, with oftentimes billions of parameters, and of such complexity that it is impossible to describe in simple terms how they work, which has led to new challenges concerning their explainability and interpretability.

# AI for Health

The recent digitalisation of all types of health information and the fact that computers are increasingly able to interpret images and text as accurately as humans[1,2] opens up countless avenues for AI applications in health. Much of the recent work on AI in health has thus gone into applications that revolve around image interpretation and natural language understanding. In the field of medical image analysis, one of the most publicized studies was by Esteva et al.[3], demonstrating the accurate classification of skin lesions using a deep neural network that was trained on clinical images, and assessing its performance by comparing its classifications to those made by board-certified dermatologists, revealing the network had reached human accuracy levels.

A survey[4] published in 2017 reviewed over 300 papers using deep learning in medical image analysis, typically for detection, segmentation, or classification tasks. The reviewed papers covered the analysis of X-ray, CT, MRI, digital pathology, cardiac, abdominal, musculoskeletal, fetal, dermatological and retinal images. In language understanding, the areas of biomedical text mining, electronic health record analysis, sentiment analysis on internet-derived data, and medical decision support systems have shown promising results[5]. Furthermore, AI methods can automatically interpret laboratory results (ranging from standard blood testing to recent advances in high-throughput genomics and proteomics) and time series (e.g. electrocardiogram, temperature, oxygen saturation, blood pressure).

A large part of the world's population has access to devices that can utilize compute-intense AI-powered applications, considering the ubiquity of computers and smartphones connected via the internet to powerful computing clusters. For example, relatively accurate detection of skin lesions using a state-of-the art camera-equipped mobile phone is technologically feasible, and medical chatbots are already on the market that can answer basic medical questions. Given the speed at which AI-based algorithms can be developed, improved, and deployed, the technology has the potential for first-class medical decision making that is accessible worldwide and affordable to the entire global population[6].

While this progress is exciting, the potential of AI for health also faces a number of challenges. In particular, deep learning models are famously hard to interpret and explain - which may substantially hinder their acceptance when facing critical or even vital decisions. Thus, interpretability, explainability, and proven robustness (e.g. to outliers and to adversarial attacks) are crucial aspects that have to be considered for trustworthiness. Moreover, health data are sensitive and subject to privacy laws. Therefore, access to sufficient training data is a major limiting factor for the predictive performance of models on data previously unseen.

This problem is complicated further because most modern AI applications are based on supervised learning and rely on data that are labeled. In the health domain, labels can typically be given only by qualified specialists, in contrast, e.g., to simple object recognition, where photographs can be labeled by legions of laypersons. In addition, machine learning approaches must take into account the biases[7] that both text and image-based medical data most likely contain. In machine learning, algorithms and training data have to be considered in combination. The algorithms can not extrapolate, but can only learn patterns that are present in the training data, which need to be of high quality, sufficiently large to learn the myriad of parameters of the "data-hungry" algorithms, and theoretically should cover all possible instances including outliers.

The **Focus Group on "Artificial Intelligence for Health"** (FG-AI4H) established by ITU in partnership with WHO aims to meet the challenges by identifying opportunities for international standardization. The initiative works on the premise that the broad adoption would benefit from a standardized and transparent evaluation of the AI methods. It should be noted that the Focus Group neither intends to specify the AI for health algorithms themselves as an ITU Recommendation, nor to standardize medical data formats, nor to establish performance criteria of hardware running the AI algorithms.

AI can support medical diagnostics and decision making by mapping from input data to output variables. Exemplary input data are images, text, time series, SNOMED CT, or HPO codes. Exemplary output variables are ICD or ICHI codes, triage tags, or other labels, depending on the use case. For instance, an AI-algorithm could map the input data to a diagnosis (represented by an ICD code), or it could make a treatment decision (represented by an ICHI code) via a diagnosis:

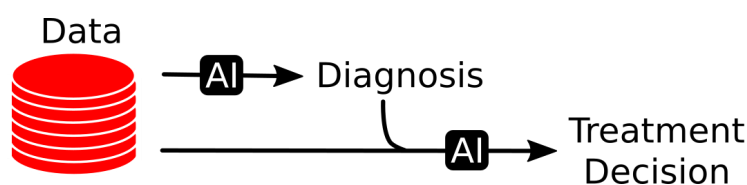

Figure 1: Mapping input data to diagnosis and treatment decision

Hence, benchmarking could be conducted without the need to disclose or standardize the AI-algorithms themselves. Instead, standardized input data sets could be created with corresponding confirmed standardized diagnosis or decision codes or other variables per patient. The data could be split into public training and private test sets. Performance metrics for comparisons could be created, which would reflect the quality of the mapping (accuracy, reproducibility, robustness, absence of bias, explainability, interpretability etc.) as well as timing aspects and other costs.

For the establishment of a benchmarking framework, it is necessary to first identify potential health problems to which AI interventions can be applied and assessed. The targeted problems and possible solutions should be scalable. Structured medical data need to be collected and open benchmarks have to be developed for the identified use cases and solutions. Subsequently the benchmarking system itself has to be implemented.

# Benchmarking Pipeline

Here, we outline a proposal for a benchmarking pipeline that will be applicable to many different scenarios. At the core of the evaluation framework lies a secret test set on which models will be evaluated.

The pipeline is summarized in the following figure:

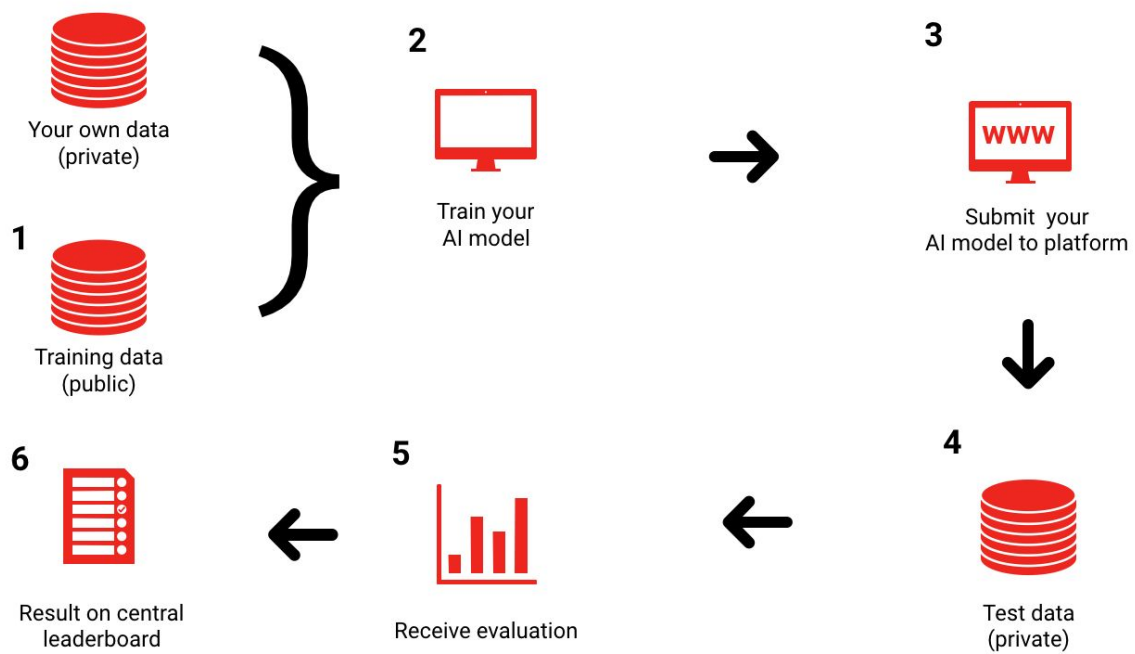

Figure 2: A Benchmarking Pipeline

The benchmarking pipeline consists of the following steps:

1. **FG-AI4H enables creation of public data repositories wherever possible**
   Most modern approaches to building AI models involve training on existing data sets. FG-AI4H will work to enable the creation of publicly available high-quality data sets to foster the creation of a diverse ecosystem of actors who want to participate in the benchmarking process.

2. **Participants build AI models based on public data and other (private) data sources**
   Participants will train their models based on a clear problem definition which is crucial for the success of a benchmark. This needs to include the quantitative measure according to which the benchmark will be assessed.

3. **Models are submitted to a benchmarking platform like crowdAI, which checks the eligibility of the model**
   Models will be submitted to agreed upon benchmarking platforms (such as [www.crowdAI.org](www.crowdAI.org)). The eligibility of the models must be defined on a case-by-case basis, but should include minimum requirements such as a maximum run time, and a maximum memory requirement.

4. **Eligible models are executed and evaluated on secret test data, managed by FG-AI4H**
   The benchmarking platform executes the model on the secret test set. The creation and governance of this secret test set will be managed by subgroups of FG-AI4H. This secret test set represents the gold standard data set for the benchmark.

5. **Participant receives evaluation, model has been benchmarked**
   After having executed the model on the secret data set, the benchmarking platform returns the evaluation results to the participant, allowing for further developments to improve the model.

6. **Central leaderboard allows comparison of model performances**
   The Benchmarking platform allows for the comparison of the models' performance on a central leaderboard, or using a pass/fail scoring. The relevant subgroups of FG-AI4H can define baselines against which models can be assessed.

It should be noted that this pipeline is meant to provide a broad overview only - developing the details, and communicating and documenting those clearly, is one of the key achievable goals of FG-AI4H and its working groups.

## Conclusion

Artificial Intelligence for Health (AI4H) offers new ways to address the shortage of medical professionals, which becomes more serious due to demographic changes and population growth. The technology has the potential to significantly improve and support medical diagnostics and treatment decision processes based on digital data. However, AI4H is rarely deployed in practice at a global scale – so far – due to legal, business, technical, or other constraints.

The new Focus Group on Artificial Intelligence for Health will work towards a standardized assessment of AI-based solutions for health, which will assure its quality, foster the adoption in practice and have a strong positive impact on global health. The FG-AI4H was founded in July 2018 by the International Telecommunication Union (ITU) in partnership with the World Health Organization (WHO), the United Nations specialized agencies for ICTs and Health, respectively. Participation in FG-AI4H is open to all researchers, engineers, practitioners, entrepreneurs and policy makers.

FG-AI4H will identify common health-specific domains (e.g. general diagnosis, specialty diagnosis [e.g. dermatology], health natural language processing, general clinical encounter note data extraction and coding, Rx coding, lab coding, etc.) and for each domain it will work for the sourcing of test data, select current gold standard test success rates (e.g. how does a professional score on this test data), set benchmark rates for AI system (to be acceptable for decision support, to be acceptable for autonomous operation), and define acceptable fail modes (e.g. alert human operator if below a given confidence threshold).

In addition to medical, scientific and technical aspects, policy, regulatory, cultural, business and other practical aspects must be considered in relationship to standardization efforts. Therefore, the FG-AI4H will establish liaisons at an early stage with selected standards bodies, forums, consortia, regulators, health professionals, core research as well as patient organizations, engineering teams, entrepreneurs and policy makers. Registries for reporting serious adverse events and guidance documents for national administrations will support the safe and appropriate use of AI in healthcare.